%% file: main.tex
\documentclass[conference]{IEEEtran}
\IEEEoverridecommandlockouts
\usepackage{cite}
\usepackage{amsmath,amssymb,amsfonts}
\usepackage{algorithmic}
\usepackage{graphicx}
\usepackage{textcomp}
\usepackage{xcolor}
\usepackage{url}
\def\BibTeX{{\rm B\kern-.05em{\sc i\kern-.025em b}\kern-.08em
    T\kern-.1667em\lower.7ex\hbox{E}\kern-.125emX}}

\usepackage{tabularx}
\newcolumntype{b}{X}
\newcolumntype{s}{>{\hsize=.45\hsize}X}
\newcolumntype{x}{>{\hsize=.25\hsize}X}
\usepackage{multirow}
\usepackage{multicol, cuted, lipsum}
\usepackage{caption}

\begin{document}

\title{Automatic Pull Request Description Generation Using LLMs: A T5 Model Approach\\
}

\author{
    \IEEEauthorblockN{Md Nazmus Sakib}
    \IEEEauthorblockA{
        \textit{\small{Department of CSE}} \\
        \textit{Pabna University of Science} \\
        \textit{and Technology} \\
        Pabna, Bangladesh \\
        \footnotesize{nazmus.200103@s.pust.ac.bd}
    }
    \and
    \IEEEauthorblockN{Md Athikul Islam}
    \IEEEauthorblockA{
        \textit{\small{Department of Computer Science}} \\
        \textit{Boise State University} \\
        Boise, ID, USA \\
        \footnotesize{mdathikulislam@u.boisestate.edu}
    }
    \and
    \IEEEauthorblockN{Md Mashrur Arifin}
    \IEEEauthorblockA{
        \textit{\small{Department of Computer Science}} \\
        \textit{Boise State University} \\
        Boise, ID, USA\\
        \footnotesize{mdmashrurarifin@u.boisestate.edu}
    }
}

\maketitle

\begin{abstract}
\input{parts/abstract}
\end{abstract}


\section{Introduction}
\label{sec:Introduction}
\input{parts/introduction}

\section{Background}
\label{sec:background}
\input{parts/background}

\section{Dataset}
\label{sec:dataset}
\input{parts/dataset}

\section{Methods}
\label{sec:methods}
\input{parts/methods}

\section{Evaluation}
\label{sec:evaluation}
\input{parts/evaluation}

\section{Implications}
\label{sec:implications}
\input{parts/implications}

\section{Conclusion and future work}
\label{sec:conclusion}
\input{parts/conclusion}

\bibliographystyle{abbrv}
\bibliography{main} 

\end{document}

%% file: parts/abstract.tex
Developers create pull request (PR) descriptions to provide an overview of their changes and explain the motivations behind them. These descriptions help reviewers and fellow developers quickly understand the updates. Despite their importance, some developers omit these descriptions. To tackle this problem, we propose an automated method for generating PR descriptions based on commit messages and source code comments. This method frames the task as a text summarization problem, for which we utilized the T5 text-to-text transfer model. We fine-tuned a pre-trained T5 model using a dataset containing 33,466 PRs. The model's effectiveness was assessed using ROUGE metrics, which are recognized for their strong alignment with human evaluations. Our findings reveal that the T5 model significantly outperforms LexRank, which served as our baseline for comparison.

%% file: parts/introduction.tex
\begin{figure}[!ht]
\centerline{\includegraphics[width=\columnwidth]{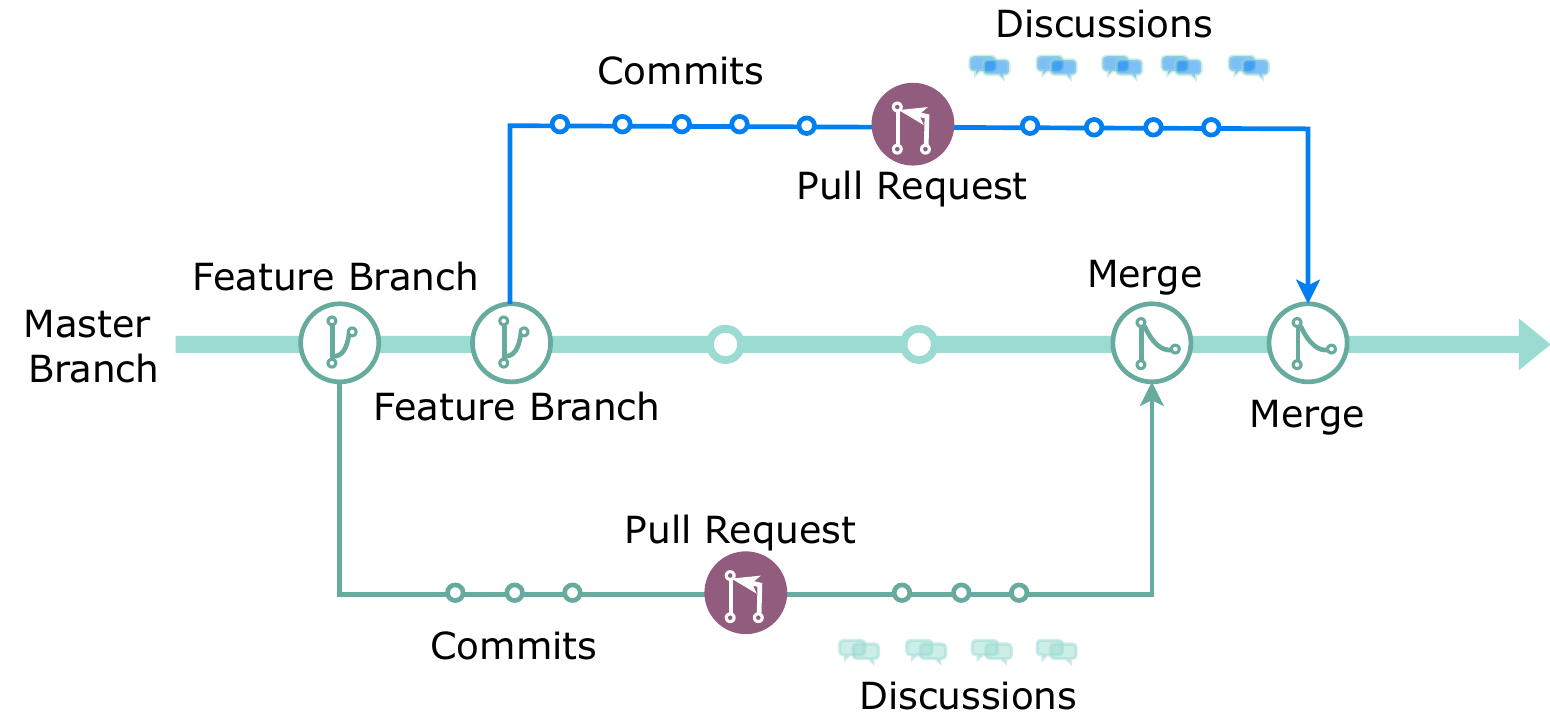}}
\caption{A typical GitHub workflow, showing the creation of feature branches from the master branch, the addition of new commits to the feature branches, and the final merge back into the master branch using PRs.}
\label{figure:github_workflow}
\end{figure}

Developers annotate their code changes with commit messages, which serve as summaries of the changes and explain the reasons behind them \cite{10.1145/1858996.1859005, yu_wait_2015}. A well-maintained commit history provides a tree view of the software's evolution over time. In contrast, a pull request (PR) or merge request combines one or more commits. During development, a milestone typically involves several commits, and upon completion, developers create a PR to request the repository owner to merge their changes. The PR undergoes a review process and discussions, and once the reviewer is satisfied, the changes are merged into the main branch \cite{10.1145/2597073.2597121}. Figure \ref{figure:github_workflow} illustrates a typical GitHub workflow, including regular commits, and the process of creating and merging PRs. When creating a PR, developers have access to all new commits and their messages that differ from the main branch. The merge process of a PR takes place after discussions and potential modifications. Since PRs provide snapshots of developers’ previous work (often called “patches”), they have become an integral part of modern software development processes, such as continuous integration \cite{7180096}.

\begin{figure}[!ht]
\centerline{\includegraphics[width=\columnwidth]{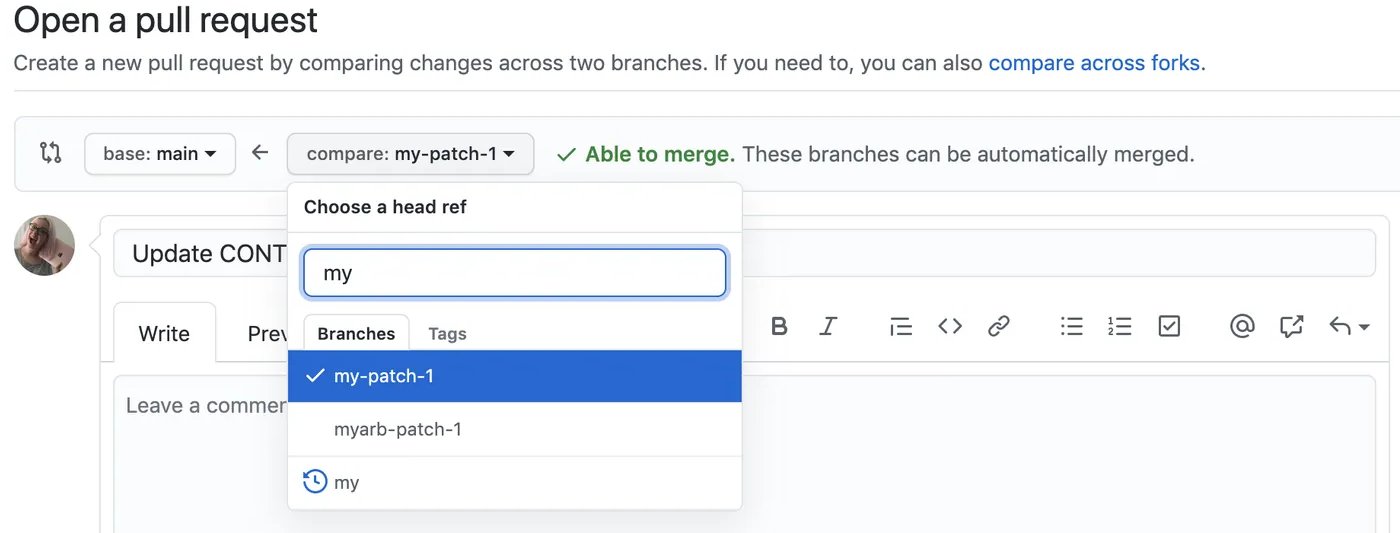}}
\caption{PR creation in GitHub \cite{githubCreatingPull}.}
\label{figure:creating_pull_request}
\end{figure}

To create a PR, developers need to provide a title and description. These are essential for explaining the purpose and details of the PR \cite{8952330}. Figure \ref{figure:creating_pull_request} illustrates the process of creating a PR for a GitHub repository \cite{githubCreatingPull}. One major benefit of a PR description is that it allows reviewers to gain an overall understanding of the PR without having to examine each commit message or the entire codebase, thereby saving a significant amount of time \cite{fan2018early, 6976151}. However, developers often neglect to include PR descriptions, either intentionally or unintentionally, leading to a substantial number of PRs with empty descriptions, which is detrimental to the project \cite{8952330}. To address this issue, it is necessary to automatically generate efficient PR descriptions.

Some researchers have proposed various approaches to generate automatic commit messages from source code. Since a PR consists of one or multiple commits, generating descriptions for these PRs requires additional care. While practitioners have attempted to generate summaries using different methods, a state-of-the-art approach to generate PR descriptions based on the T5 transformer model is lacking. Therefore, we propose an automatic PR description generation method using a large language model (LLM) such as the unified text-to-text transfer transformer model T5. By leveraging this technique, teams can easily populate all empty PR descriptions using existing commit messages.

%% file: parts/background.tex
Researchers have proposed numerous approaches to automate software changes, including the generation of commit messages. One notable example is CommitBERT, proposed by Tae-Hwan Jung, which summarizes code changes and addresses the challenge developers face in writing commit messages \cite{DBLP:journals/corr/abs-2105-14242}. Another tool, ChangeScribe, automatically generates commit messages by summarizing source code changes \cite{7203049}. Additionally, practitioners have utilized translation-based machine learning techniques to generate commit messages from git diffs \cite{8115626}. These approaches typically employ natural language processing (NLP) techniques, using datasets that comprise code changes and commit messages to train their models and automate commit message generation.

In addition to commit messages, practitioners have also explored automating the process of generating descriptions for PRs. Fang et al. aimed to enhance the efficiency of PR description generation models \cite{FANG2022111160}. They first identified the limitations of RNN-based models and proposed a novel approach called a hybrid attention network, which outperformed RNN-based models in terms of speed and effectiveness. Kuang et al. focused on addressing large-granularity PRs \cite{15430530020211001}. They introduced a model that captures extensive information about PR sentences to establish connections between them. Liu et al. treated the PR description generation problem as a sequence-to-sequence learning task and developed an innovative encoder-decoder model \cite{8952330}. This model addresses out-of-vocabulary words and enhances ROUGE optimization by incorporating a specialized loss function.

To tackle the challenge of description generation, researchers have explored various NLP and machine learning techniques. Among the most commonly utilized methods are the Seq2Seq model, attention network, and global attention network \cite{FANG2022111160}. Transformer-based models, which utilize the self-attention mechanism, have emerged as preferred choices for many natural language generation tasks, including machine translation and text summarization \cite{NIPS2017_3f5ee243, you-etal-2019-improving}. BERT (Bidirectional Encoder Representations from Transformers) is a prominent example of such models, capable of pretraining deep bidirectional representations \cite{devlin2018bert}. These pre-trained representations can be fine-tuned by adding an extra output layer \cite{devlin2018bert}. One significant advantage of fine-tuning these pre-trained models is their versatility across various language-related tasks, such as question answering, summarization, and language interface, without substantial architectural changes \cite{devlin2018bert}.

Following the success of BERT, Google introduced the Text-To-Text Transfer Transformer, or T5, a unified model architecture capable of performing various text-to-text tasks \cite{https://doi.org/10.48550/arxiv.1910.10683}. While BERT excels at tasks like classification or span prediction, where it outputs labels or spans corresponding to input sentences, T5 transformers are particularly effective for generative tasks such as translation or abstractive summarization \cite{https://doi.org/10.48550/arxiv.1910.10683}. Unlike BERT, which is not well-suited for these types of tasks, T5 demonstrates superior performance in such scenarios.

Practitioners have already employed various strategies, including the use of a hybrid attention network and BERT, to automate the process of generating PR descriptions. However, the approach of fine-tuning a widely-used pre-trained unified text-to-text transfer transformer model, such as T5, for this purpose, along with its evaluation, is lacking in the current state of the art. Our plan is to fine-tune a pre-trained T5-small model and evaluate its performance using ROUGE metrics.

We pose the following research question to drive the research -

\begin{itemize}
    \item \textit{RQ: Is the T5 model effective for automatic PR description generation?}
\end{itemize}

%% file: parts/dataset.tex
\subsection{Data Collection}

We utilized a dataset sourced from GitHub, containing 33,466 filtered PRs selected from a total of 333,000 PRs \cite{8952330}. Table \ref{table:pull_request_des_com} displays a sample PR from the training dataset, including its ID, description, and associated commit messages. The data was gathered from 95,804 Java repositories, of which 22,700 had at least one merged PR \cite{8952330}. These repositories were sorted in descending order by the number of merged PRs, and the dataset was compiled using GitHub’s APIs \cite{8952330}. The input data consists of a combination of commit messages and source code comments, while the output is the PR description. Table \ref{table:datasets} outlines the dataset split: 80\% for training, 10\% for validation, and 10\% for testing.

\begin{table}[h!]
\centering
\def\arraystretch{1.3}
\small
\setlength{\tabcolsep}{4pt}
\resizebox{\columnwidth}{!}{
\begin{tabular}{>{\raggedright\arraybackslash}p{2.5cm} >{\raggedright\arraybackslash}p{2.5cm} >{\raggedright\arraybackslash}p{2.5cm}}

\hline
\hline
Split & \# of examples & Language \\
\hline
Train & 26,772 & JAVA \\
Validation & 3,347 & JAVA \\
Test & 3,347 & JAVA \\
\hline
\hline
\end{tabular}
}

\caption{Statistics of Dataset}
\label{table:datasets}
\end{table}

\subsection{Data Preprocessing}

The first preprocessing step involved removing HTML or rich-text elements from both the commit messages and PR descriptions. Additionally, all URLs, references (e.g., "\#456"), signatures (e.g., "reviewed-by"), email addresses (e.g., "@placeholder"), and markdown headlines were eliminated using the NLTK library \cite{8952330}. In the datasets, comments associated with a specific PR were separated by the '{\textless cm-sep\textgreater}' separator. 
We replaced all instances of this separator to ensure that the model does not include them in the learning process. Subsequently, we created two bar charts to visualize the length distribution of abstracts (PR descriptions) and articles (commit messages). We also calculated the average length of articles and abstracts, which were found to be 76.02 and 36.41, respectively.

\begin{table*}[htbp]
\begin{center}
\def\arraystretch{1.5}
\begin{tabularx}{\textwidth}{b b}
\hline
\hline
\textbf{PR Description:} a new plugin has been declared maven-java-formatter-plugin . to use it , use ' mvn java-formatter : format ' . this plugin stands upon a new esigate module : esigate-tool-conf which should be used in the future to hold pmd ( already present in the new module ) and checkstyle ( present at project root level ) configuration . ( code formatting has not been made )  \newline
\textbf{ID:} esigate/esigate\_11 
& 
\textbf{Commit 1:} \newline
\textbf{Commit Message:} fix git ignore multiplicated settings. \newline
\textbf{Commit 2:} \newline
\textbf{Commit Message:}  change path to formater config file. \newline
\textbf{Commit 3:} \newline
\textbf{Commit Message:} \newline format plugin attached to compile and defined in each module that needs format .\newline
\textbf{Commit 4:} \newline
\textbf{Commit Message:} \newline checkstyle configured the right way for multimodule project and complete makeover for site generation. \\
\hline
\hline
\end{tabularx}
\end{center}
\caption{The left column contains the pull request descriptions and IDs. The right column lists the four commit messages from which the PR was created.}
\label{table:pull_request_des_com}
\end{table*}

%% file: parts/methods.tex
\begin{figure}[htbp]
\centerline{\includegraphics[width=\columnwidth]{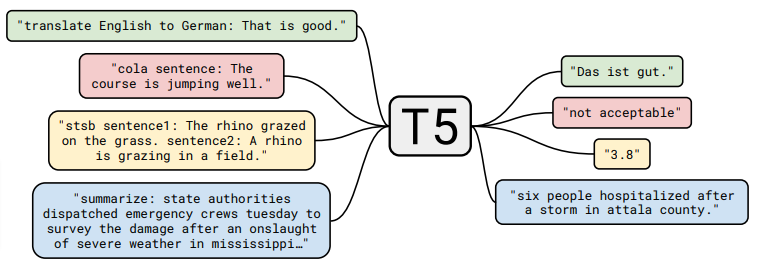}}
\caption{Text-to-text framework \cite{https://doi.org/10.48550/arxiv.1910.10683}}
\label{figure:text_to_text_framework}
\end{figure}

To evaluate the effectiveness of the T5 transformer for PR description generation, we approach this text processing problem as a "text-to-text" task \cite{https://doi.org/10.48550/arxiv.1910.10683}. This means that we take text as input and generate another text as output. Figure \ref{figure:text_to_text_framework} illustrates how the text-to-text framework can be applied to tasks such as translation, sentiment analysis, and, most importantly, summarization in our case.

We leveraged transfer learning by utilizing the pre-trained Hugging Face T5 (t5-base) transformer, which incorporates a multi-task mixture of unsupervised and supervised tasks. The T5 model was pre-trained on the publicly available Colossal Clean Crawled Corpus (C4) dataset \cite{https://doi.org/10.48550/arxiv.1910.10683}. Subsequently, we fine-tuned the model using our smaller labeled datasets, aiming to achieve significantly improved performance compared to training the model solely on our data.

Our model construction and approach involve the following steps:

\subsection{Tokenization}

\begin{figure}[htbp]
\centerline{\includegraphics[width=\columnwidth]{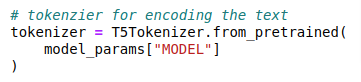}}
\caption{T5 Tokenizer}
\label{figure:tokenizer}
\end{figure}

Tokenization is the initial step in processing inputs, where text is split into tokens before being passed into the model. In the final stage, tokens are converted back into words. The T5 tokenizer uses the SentencePiece approach, which tokenizes and detokenizes text in an unsupervised manner. Figure \ref{figure:tokenizer} illustrates how we instantiated the tokenizer. Additionally, we utilized the batch\_encode\_plus method of the tokenizer to construct the source and target. These dictionaries contain encoded sequences or sequence pairs along with additional information. The max\_length parameter is employed to limit the total sequence length.

\subsection{Constructing Target Sequence}

The target sequence of the PRs comprises their descriptions. These target sequences are constrained to a maximum length of 50 tokens, which includes both words and punctuation. After calculating the average length of the abstracts as 36.41 tokens, we opted to set the maximum length to 50 tokens. Additionally, we enabled the truncation parameter to be true to ensure that the model can handle longer sequences.

\subsection{Constructing Source Sequence}

The source sequence of the PRs consists of both the commit messages and source code comments. We set a maximum length of 512 tokens for the source sequences, which includes both words and punctuation. Similar to the target sequence, the source sequence underwent direct preprocessing using a standard text preprocessing procedure.

\subsection{Trainer model}

We developed a T5-trainer to fine-tune the T5 model using our labeled source and target data. This T5-trainer first initializes the tokenizer instance from the tokenizer class and then adds a language model layer on top to generate the target sequence. It processes the source and target texts, constructing a data frame from them and generating the corresponding sequences using the tokenizer. Subsequently, it utilizes PyTorch DataLoader to create separate train and test data loaders. For optimization, the Adam optimizer from PyTorch is employed to adjust the network weights during training. The T5-trainer also specifies the number of epochs and other hyperparameters, training the model accordingly.

%% file: parts/evaluation.tex
For the evaluation, we employed ROUGE metrics. In this section, we will describe the ROUGE metrics utilized in our evaluation and discuss how they compare to the baseline.

\subsubsection{ROUGE metrics}

\begin{equation}
\label{eq:rouge_n}
ROUGE-N = \frac{\sum_{S \in \text{Reference Summaries}} \sum_{gram_n \in S} \text{Count}_{\text{match}}(gram_n)}{\sum_{S \in \text{Reference Summaries}} \sum_{gram_n \in S} \text{Count}(gram_n)} 
\end{equation}

\begin{align}
R_{LCS} &= \frac{LCS(X, Y)}{|Y|} \label{eq:rouge_l_r} \\[1.4em]
P_{LCS} &= \frac{LCS(X, Y)}{|X|} \label{eq:rouge_l_p} \\[1.4em]
F_{LCS} &= \frac{(1 + \beta^2) \cdot R_{LCS} \cdot P_{LCS}}{R_{LCS} + \beta^2 \cdot P_{LCS}} \label{eq:rouge_l_f}
\end{align}

\hspace{1em}

ROUGE (Recall-Oriented Understudy for Gisting Evaluation) assesses the quality of summaries by comparing them to manually created references \cite{6976151}. In our evaluation, we utilized ROUGE-N (ROUGE-1 and ROUGE-2) and ROUGE-L metrics. Equation \ref{eq:rouge_n} demonstrates the calculation process for ROUGE-N. Here, the subscript 'n' represents the length of the n-gram, and ROUGE-N is a recall-based measure as the ratio of the number of matched n-grams to the total number of n-grams in the reference summaries.

On the other hand, ROUGE-L is based on LCS (Longest Common Subsequence). Equations \ref{eq:rouge_l_r}, \ref{eq:rouge_l_p}, and \ref{eq:rouge_l_f} illustrate the calculations for recall, precision, and F1 score for ROUGE-L, respectively. The recall-based LCS ($R_{LCS}$) is the ratio of the length of the LCS to the length of the reference summary. The precision-based LCS ($P_{LCS}$) is the ratio of the length of the LCS to the length of the candidate summary. The F1 score based on LCS ($F_{LCS}$) is the harmonic mean of $R_{LCS}$ and $P_{LCS}$, weighted by $\beta$. They identify the longest subsequence between the output and target from the test dataset. We evaluated both ROUGE-N and ROUGE-L metrics using Python's ROUGE package.

\subsubsection{Baseline}

To establish a baseline for PR description generation, we utilized LexRank, which is superior to centroid-based methods \cite{erkan2004lexrank}. LexRank constructs clusters using sentences from documents and then selects the central sentences that reflect the summarized idea of each cluster. It employs eigenvector centrality within a graph representation of sentences to determine the importance of word tokens.

\subsubsection{RQ: Is the T5 model effective for automatic PR description generation?} 

To evaluate our research question, we initially trained the model using the maximum batch size that the GPU memory supports (batch\_size=16) for 5 epochs. Subsequently, during evaluation, we used the fine-tuned T5-trainer model to generate predictions for the test split. We saved these predictions, along with the actual PR descriptions, to a file named 'predictions.csv' within the 'model\_files' directory. Our model validation process begins with validating the 'predictions.csv' file. This involves calculating the ROUGE-1, ROUGE-2, and ROUGE-L metrics based on the predicted and original PR descriptions. We then compare these metric values to those obtained using the baseline LexRank.

We present our evaluation results in Table \ref{table:evaluation_comparison}, comparing our approach (T5) with the baseline approach (LexRank). The F1 scores produced by T5 are 32.16, 20.26, and 28.92 for ROUGE-1, ROUGE-2, and ROUGE-L, respectively. These scores align with the standard F1 scores for summarization tasks, which typically range from 0.20 to 0.40 \cite{https://doi.org/10.48550/arxiv.1704.04368, https://doi.org/10.48550/arxiv.1705.04304}. Furthermore, our T5 approach outperforms the baseline LexRank across all metrics, with average absolute gains of 4.35, 13.92, and 6.65 for recall, precision, and F1 score, respectively. These results indicate that our model can accurately capture the main points of PR descriptions compared to LexRank.

One reason T5 excels at producing high-quality PR descriptions is its training on the diverse "Colossal Clean Crawled Corpus" (C4), a cleaned and structured version of the Common Crawl web dump. The C4 dataset, which encompasses 250 billion pages collected over the past 17 years enhances the model's ability to understand and generate natural language text \cite{commoncrawl}. In addition to this extensive pre-training, T5 leverages the inherent strengths of the Transformer architecture, particularly its self-attention mechanism. This self-attention mechanism allows T5 to excel in text-to-text tasks by effectively capturing long-range dependencies within the text \cite{cheng-etal-2016-long}. This ability to maintain context and coherence across longer passages is crucial for producing accurate and relevant PR descriptions. Furthermore, the benefits of transfer learning are clearly demonstrated in our use case. While our specific dataset of PR descriptions was relatively small, fine-tuning the pre-trained T5 model on this dataset allowed us to harness the extensive knowledge encoded in the large C4 dataset. Overall, the combination of T5's training on a diverse and extensive corpus, its advanced Transformer-based architecture, and the effective application of transfer learning contributes to its superior performance in generating high-quality PR descriptions.

\begin{table*}[htbp]
\begin{center}
\def\arraystretch{1.4}
\begin{tabularx}{\textwidth}{l | x x x | x x x | x x x}

\hline
\hline

\multirow{2}{*}{Approach} & \multicolumn{3}{c|}{ROUGE-1} & \multicolumn{3}{c|}{ROUGE-2} & \multicolumn{3}{c}{ROUGE-L}   \\ 

\cline{2-10} 

& Recall & Precision & F1 score & Recall & Precision & F1 score & Recall & Precision & F1 score \\ 
\hline
LexRank & 25.19  & 30.11 & 25.30 & 12.69 & 13.81     & 12.26 & 25.28 & 28.33 & 23.82 \\ 
\hline
T5 & \textbf{29.68}  & \textbf{46.89} & \textbf{32.16} & \textbf{19.56} & \textbf{26.08} & \textbf{20.26} & \textbf{26.96} & \textbf{41.03} & \textbf{28.92} \\ 
\hline

LexRank vs T5 & +17.82\% & +55.73\% & +27.11\% & +54.14\% & +88.85\% & +65.26\% & +6.65\% & +44.83\% & +21.41\%  \\ 

\hline
\hline

\end{tabularx}
\end{center}
\caption{Comparison of T5 and baseline model LexRank. The best performance is marked in \textbf{bold}.}
\label{table:evaluation_comparison}
\end{table*} 

\begin{table}[]
\begin{center}
\def\arraystretch{1.1}
\begin{tabular}{|l|}
\hline
\begin{tabular}[c]{@{}l@{}}In summary,  the T5 model outperforms the\\ baseline LexRank in terms of ROUGE-1, \\ ROUGE-2, and ROUGE-L, and produces \\ more precise descriptions.\end{tabular} \\ \hline
\end{tabular}
\end{center}
\end{table}

%% file: parts/implications.tex
Our model is designed to assist developers who may forget to write PR descriptions or encounter difficulties in crafting meaningful ones. For instance, during the development of a PR that spans several days, developers may overlook key points from individual commits. In such cases, our PR summaries serve as valuable reminders. Additionally, developers can compare our generated summaries with their handwritten descriptions to identify any missed points.

Furthermore, our model aids reviewers in swiftly grasping the main ideas of PRs without the need to sift through each commit message and code change. This efficiency enables quicker decision-making, saving time in the review process. Moreover, automated PR descriptions can facilitate ongoing software maintenance efforts.

%% file: parts/conclusion.tex
In this paper, our objective was to automate the generation of PR descriptions from commit messages and assess the effectiveness of our approach. We framed the PR description generation task as a sequence-to-sequence problem and employed a text-to-text T5 transformer. Leveraging transfer learning, we utilized a pre-trained T5 transformer and fine-tuned it on our smaller dataset comprising 10k PRs. Evaluation using ROUGE metrics demonstrated that our comprehensive model surpassed the baseline LexRank model in performance.

Moving forward, our future endeavors include training different LLMs, also the models on larger datasets and taking into account environmental constraints encountered during training on extensive datasets. Additionally, we aim to investigate whether our model maintains the same level of accuracy for other language generation tasks. Furthermore, we intend to extend the application of our model to other software engineering tasks, such as source code summarization and bug report summarization.